\pdfoutput=1

\documentclass[11pt]{article}

\usepackage{ACL2023}

\usepackage{times}
\usepackage{latexsym}

\usepackage[T1]{fontenc}

\usepackage[utf8]{inputenc}

\usepackage{microtype}

\usepackage{inconsolata}

\usepackage{microtype}
\usepackage{amsmath}
\usepackage{amssymb}
\usepackage{graphicx}
\usepackage{tikz}
\usepackage{pgf}
\usepackage{xcolor}
\usepackage{booktabs, makecell}
\usepackage{enumitem}
\usepackage{subcaption}
\usepackage{array}

\usepackage{contour}
\usepackage[normalem]{ulem}

\contourlength{1pt}
\newcommand{\unline}[1]{%
  \uline{\phantom{#1}}%
  \llap{\contour{white}{#1}}%
}
\newcommand{\dashunline}[1]{%
  \dashuline{\phantom{#1}}%
  \llap{\contour{white}{#1}}%
}

\newcommand{\relation}[1]{\texttt{\small #1}}
\newcommand{\predicatetext}[1]{\textcolor{violet}{\textbf{#1}}}

\title{Smoothing Entailment Graphs with Language Models}

\author{Nick M\raisebox{0.3ex}{c}Kenna$^\dagger$ \quad Tianyi Li$^\dagger$ \quad Mark Johnson$^\ddagger$ \quad Mark Steedman$^\dagger$ \\
    $^\dagger$University of Edinburgh \quad $^\ddagger$Macquarie University \\
    \texttt{nick.mckenna@ed.ac.uk} \quad \texttt{tianyi.li@ed.ac.uk} \\ \texttt{mark.johnson@mq.edu.au} \quad
    \texttt{steedman@inf.ed.ac.uk}
  }

\begin{document}
\maketitle

\begin{abstract}
The diversity and Zipfian frequency distribution of natural language predicates in corpora leads to sparsity in Entailment Graphs (EGs) built by Open Relation Extraction (ORE). EGs are computationally efficient and explainable models of natural language inference, but as symbolic models, they fail if a novel premise or hypothesis vertex is missing at test-time. We present theory and methodology for overcoming such sparsity in symbolic models. First, we introduce a theory of optimal \textit{smoothing} of EGs by constructing transitive chains. We then demonstrate an efficient, open-domain, and unsupervised smoothing method using an off-the-shelf Language Model to find approximations of missing premise predicates. This improves recall by 25.1 and 16.3 percentage points on two difficult directional entailment datasets, while raising average precision and maintaining model explainability. Further, in a QA task we show that EG smoothing is most useful for answering questions with lesser supporting text, where missing premise predicates are more costly. Finally, controlled experiments with WordNet confirm our theory and show that hypothesis smoothing is difficult, but possible in principle.\footnote{Code available at \href{https://github.com/nighttime/EntGraph}{github.com/nighttime/EntGraph}}
\end{abstract}

\section{Introduction}

\begin{figure}[t]
    \centering
    \includegraphics[width=0.97\linewidth]{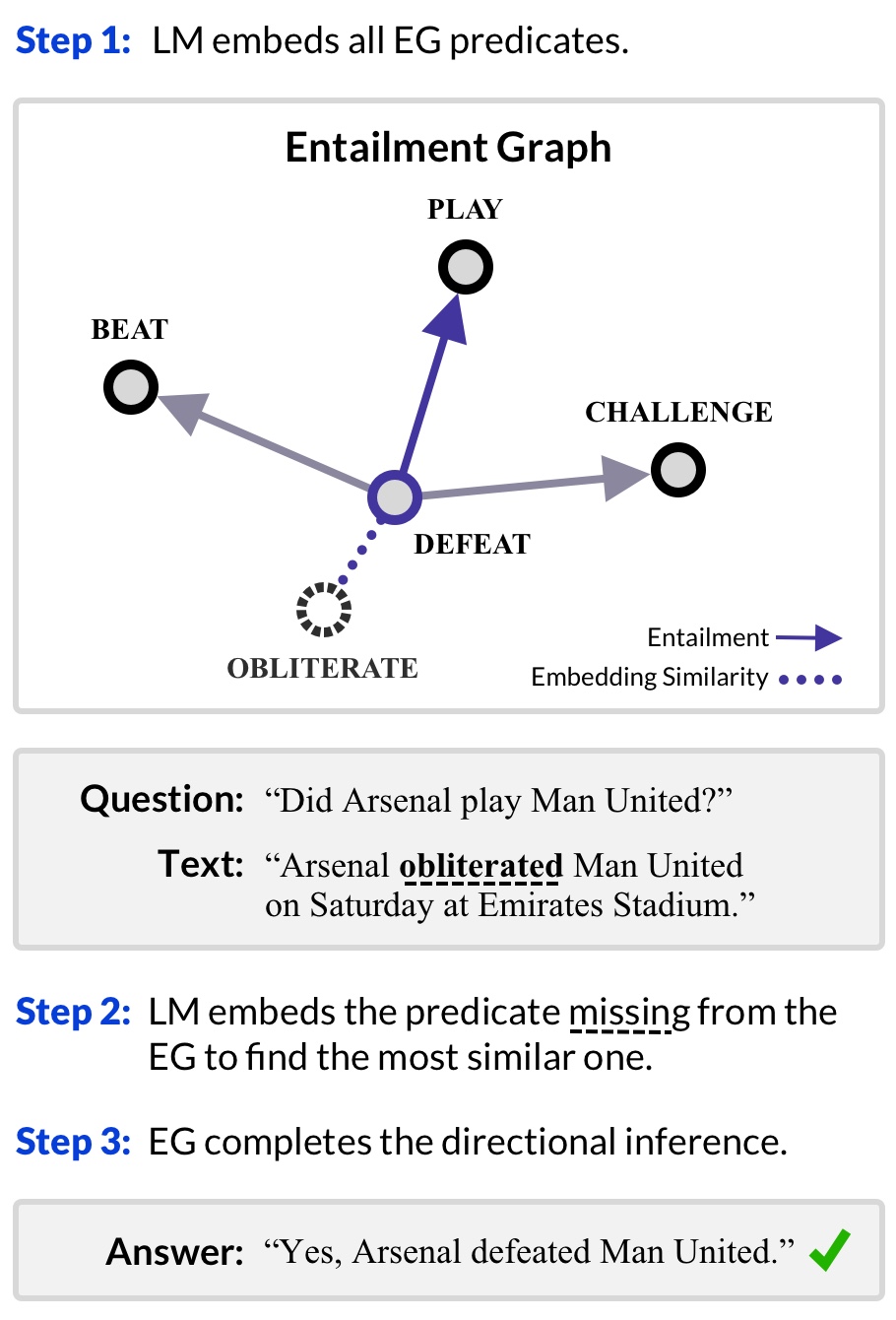}
    \caption{The question cannot be answered because a predicate in the text isn't in the Entailment Graph. An LM embeds the predicate so a nearest neighbor in the EG can be found, completing the directional inference.}
    \label{fig:method-illustration}
    \vspace{-0.5cm}
\end{figure}

An Entailment Graph (EG) is a learned structure for making natural language inferences of the form [premise] \textit{entails} [hypothesis], such as ``if Arsenal \textbf{defeated} Man United, then Arsenal \textbf{played} Man United.'' An EG consists of a set of vertices (typed natural language predicates), and a set of directed edges constituting entailments between predicates. They are constructed in an unsupervised manner using the Distributional Inclusion Hypothesis \cite{geffet-dagan-2005-distributional}: a representation is generated for each predicate based on its distribution with arguments in a training corpus, and representation subsumption is used for learning directional entailments between predicates. A \textit{directional inference} is stricter than paraphrase or similarity, in that it is true only in one direction, but not both, e.g. $\textsc{defeat} \vDash \textsc{play}$ but $\textsc{play} \nvDash \textsc{defeat}$ (where $\vDash$ means ``entails''). Directional inferences are difficult to learn, but crucial to language understanding.

EGs are useful in tasks like Knowledge Graph link prediction \cite{hosseini2019-duality, hosseini-etal-2021-open-domain} and question answering from text \cite{lewis_semantics_2013, mckenna-etal-2021-multivalent}. EG learning is unsupervised: building them only requires a parser and entity linker for a new language domain \cite{li-etal-2022-cross-lingual}. EGs are relatively very data- and compute-efficient, requiring less than two days to train on 2GB of unlabeled text using a single GPU \cite{hosseini-etal-2021-open-domain}. Further, EGs are editable and also explainable, because decisions can be traced back to distinct sentences on a task.

However, EGs suffer from two kinds of sparsity. One is \textit{edge sparsity}, when two predicates are not observed with co-occurring entities, so cannot be connected together. Recent work improves on EG connectivity \cite{berant-etal-2015-efficient, hosseini2021unsupervised, chen-etal-2022-transitivity} but to our knowledge we are the first to acknowledge \textit{vertex sparsity}, arising when a predicate is not seen at all in training. EGs are structures of symbols, so they cannot handle missing queries: in an inference task, if \textit{either} the premise or hypothesis predicate is not in training, no entailment edge can be learned. In fact, many EG demonstrations achieve just 50\% of task recall. Predicates occur in a Zipfian frequency distribution with an unbounded tail of rare predicates, so it's impractical to scale up the learning of predicate symbols by reading larger corpora. There will virtually always be predicates missing at test-time.

Modern Language Models combine representations of subword tokens to solve a similar issue \cite{peters-etal-2018-deep, devlin-etal-2019-bert}, and recent scaling of LMs has lead to breakthrough performance on many tasks \cite{chinchilla-hoffman-2022, wei2022emergent}, offering relief to sparsity problems via techniques like in-context learning \cite{gpt3-brown-2020}. However, as LMs scale in size and compute they bring new problems: they require ballooning GPU resources to train or run; or are costly to query via API; and centralizing models under private companies opens challenges of data privacy. We are thus motivated to research lower-compute and more data-efficient methods which run on the scale of a single GPU. 

We are the first to define vertex sparsity and approach the problem by applying a small, pretrained LM to improve an existing EG using the benefits of modern embeddings. We offer four contributions:



(1) A theory for optimal smoothing of symbolic inference models such as EGs by constructing transitive chains, accounting for a distinction between premise and hypothesis.

(2) A low-compute method for unsupervised smoothing of EG vertices using LM embeddings to find approximations of missing predicates (see Figure \ref{fig:method-illustration}). Applied to premises, we improve recall by 16.3 and 25.1 percentage points on Levy/Holt and ANT entailment datasets while raising precision. 

(3) On a QA task we show LM premise smoothing is most helpful when there is less supporting context and missing a predicate is more costly.

(4) Finally, in controlled experiments with WordNet relations we confirm the behavior of the LM for premise smoothing and show that hypothesis smoothing is possible, but more difficult.

\section{Background}
\label{sec:background}

Research on unsupervised Entailment Graph induction has mainly oriented toward edges: overcoming edge sparsity using graph properties like transitivity \cite{berant-etal-2015-efficient, hosseini2018, chen-etal-2022-transitivity}, incorporating contextual or extralinguistic information to improve edge precision \cite{hosseini-etal-2021-open-domain, guillou-etal-2020-incorporating}, and research into the underlying theory of the Distributional Inclusion Hypothesis \cite{kartsaklis-sadrzadeh-2016-distributional, mckenna-etal-2021-multivalent}. However, none of these address vertex sparsity.

 
We leverage sub-symbolic encoding by an LM using WordPieces \cite{devlin-etal-2019-bert} in this work as a means of smoothing, to generalize beyond a fixed vocabulary of predicates. Our most direct comparison is with \citet{schmitt-schutze-2021-language} who apply contemporary prompting techniques with the computationally tractable RoBERTa \cite{liu2019roberta} to learn open-domain predicate entailment. They finetune on premise-hypothesis pairs and labels from the development split of the Levy/Holt NLI dataset \cite{holt_2018}, used in our experiments. They use templates like ``[hypothesis], because [premise]'' which are encoded by the LM, then classified true/false. They report high scores on datasets, but \citet{li-lm-poor-learner} have shown that despite excelling at paraphrase detection, rather than learning directional inference (e.g. \textsc{buy} $\vDash$ \textsc{own} and \textsc{own} $\nvDash$ \textsc{buy}), this technique picks up dataset artifacts spuriously correlated with the labels in Levy/Holt. In contrast, our approach combines the strengths of each: open-domain encoding using a computationally tractable LM with the directional inference capability of an EG.

\section{Theory of Smoothing}

We first present a theory for optimal smoothing of a symbolic EG which overcomes the problem of vertex sparsity. We define \textit{smoothing} as the approximation of missing predicates using those in the existing predicate vocabulary, in reference to earlier work in smoothing n-gram Language Models \cite{chen-goodman-1996-empirical}. We next discuss the theoretical intuition behind applying an LM as an open-domain smoother.


\subsection{Directionality by Transitive Chaining}
\label{sec:transitive_chaining}

We argue that it is most important when modifying EG predictions by smoothing to maintain the EG's strong directional inference capability. Our theory maintains directionality by constructing transitive chains, importantly distinguishing the \textit{role} of the proposition as premise or hypothesis. We distinguish ways to \textbf{P-smooth} premises and \textbf{H-smooth} hypotheses.

We start with a query entailment relation $Q: p \vDash h$, with unknown truth value to be verified by a model which is \dashunline{missing} entries for at least $p$ or $h$. We specify smoothing as the process of generating a new relation $Q_s$ suitable for the model by identifying a \unline{replacement} predicate $p'$ and/or $h'$ within the model's vocabulary. We claim that to maintain directional precision, this must be done by identifying a $p'$ (or $h'$) related to $p$ (or $h$) such that a transitive chain is constructed, as in the cases below. By this transitivity, confirmation of $Q_s$ is leveraged to confirm $Q$.

\begin{enumerate}[leftmargin=*]
    \item \textbf{Generalize Missing P.} Identify a more general premise $p'$ in the EG such that $p \vDash p'$. This yields a new $Q_s: p' \vDash h$. \\[0.25cm]
    \begin{tabular}{lccc}
        ($Q$) & \text{``$a$ \dashunline{obliterated} $b$''} & $\vDash$ & \text{``$a$ played $b$''} \\
        & \rotatebox[origin=c]{270}{$\vDash$} & & \\
        ($Q_s$) & \text{``$a$ \unline{{beat}} $b$''} & $\vDash$ & \text{``$a$ played $b$''}
    \end{tabular} 
    
    $p \vDash p'$ is known, so if the EG confirms $p' \vDash h$, then $p \vDash h$ is confirmed by transitivity.
    
    \item \textbf{Specialize Missing H.} Identify a more specialized hypothesis $h'$ in the EG such that $h' \vDash h$. This yields a new $Q_s: p \vDash h'$. \\[0.25cm]
    \begin{tabular}{lccc}
        ($Q$) & \text{``$a$ bought $b$''} & $\vDash$ & \text{``$a$ \dashunline{shopped for} $b$''} \\
        & & & \rotatebox[origin=c]{90}{$\vDash$} \\
        ($Q_s$) & \text{``$a$ bought $b$''} & $\vDash$ & 
        \text{``$a$ \unline{{paid for}} $b$''}
    \end{tabular} 
    
    If the EG confirms $p \vDash h'$, then also knowing $h' \vDash h$ confirms $p \vDash h$ by transitivity.
    
    \item \textbf{Generalize P and Specialize H.} If missing both $p$ and $h$, combine methods: identify new $p'$ and $h'$ as above, yielding a new $Q_s: p' \vDash h'$. 
    
    Knowing $p \vDash p'$ and $h' \vDash h$, if a model confirms $p' \vDash h'$, then $p \vDash h$ is confirmed by transitivity.
\end{enumerate}








Of course, the success of this smoothing depends on being able to find $p'$ such that $p \vDash p'$, and $h'$ such that $h' \vDash h$. However, when an additional inference is found, it is likely to be correct, aiding model precision. By definition we cannot use the EG for this, and we turn to Language Models to identify replacement predicates.


\subsection{LM Embeddings and Specificity}
\label{sec:discussion_of_asymmetry}

We assume that $p'$ and $h'$ are respectively among the nearest neighbors of p and h in the embedding space of the LM, and in this paper propose a method to leverage LM embeddings in an unsupervised way to find them. As defined later in \S\ref{sec:smoothLM}, we first embed all EG predicates, then at test-time we embed the target query predicate and search for the K nearest neighbors to the target in embedding space. We predict that doing so for a premise predicate will build a transitive chain satisfying the conditions of \S\ref{sec:transitive_chaining}. We identify two factors which, combined, lead to predictions that are likely more semantically general than the target, which enables P-smoothing, but not H-smoothing:



\textbf {(A)} The LM training objective. \citet{li-etal-2020-sentence} show that the masked language modeling objective in BERT induces a particular structure in its latent embedding space: on average, corpus-frequent words are embedded near the origin and infrequent ones further out. This is because of statistical learning, which biases LMs toward high frequency words since they are trained on a corpus to predict the most probable tokens. This objective leads LSTM-based LMs to produce a beneficially Zipfian frequency distribution of words \cite{takahashi2017zipf}, and similar biases are evident in Transformers for generation like GPT-2 and XLNet \cite{shwartz-choi-2020-neural}.

\textbf{(B)} The natural anti-correlation of word frequency with specificity in text. Probabilistically, the more frequent a word, the lower its ``semantic content'' (in other words, the less specific it is). \citet{caraballo-charniak-1999-determining} show this for nouns, and this assumption is even used in the ``IDF'' component of TF-IDF \cite{jones1972statistical}.

These factors imply that embedding a vocabulary of EG predicates using an LM will result in a space densely populated toward the origin by corpus-frequent predicates. KNN-search starting from a target predicate embedding will likely return neighbors toward this dense origin, thus selecting more corpus-frequent, semantically general words. We illustrate further in \S\ref{sec:specificity_taxonomy}. 

This effect has even been studied elsewhere: in Machine Translation, frequency bias causes a quantified semantic generalizing effect from translation input to output \cite{vanmassenhove-etal-2021-machine}, dubbed ``Machine Translationese'' due to the artificially non-specific tone.


\subsection{The Specificity Taxonomy}
\label{sec:specificity_taxonomy}

\begin{figure}[h]
\centering
    \includegraphics[width=0.95\linewidth]{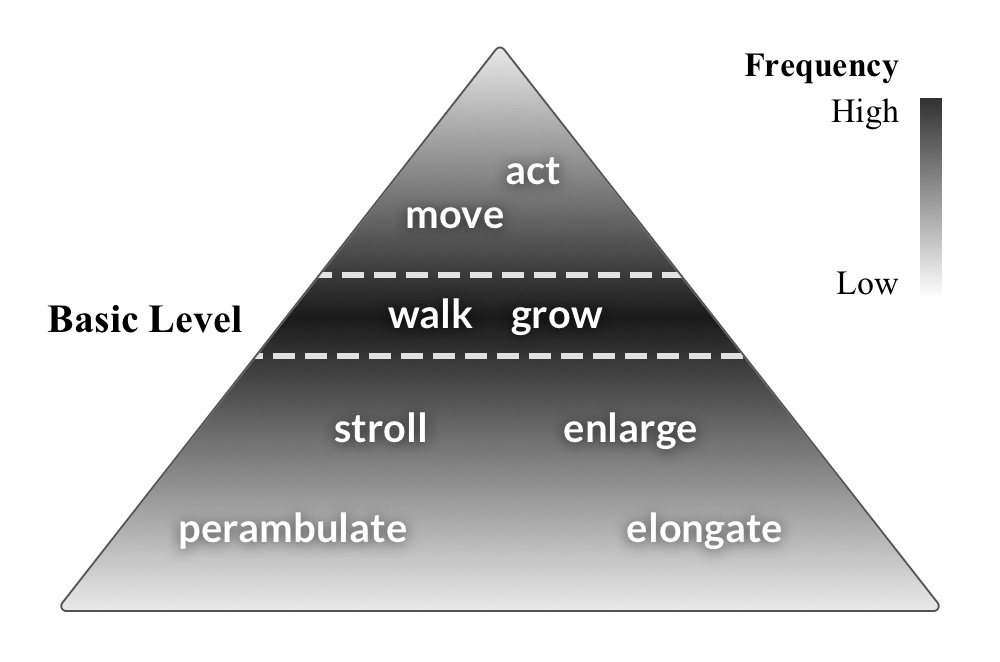}
    \vspace{-0.25cm}
    \caption{The specificity taxonomy. The basic level contains ``everyday'' predicates. Above becomes more general, and below becomes more concrete and specific. Usage frequency decreases away from the basic level.}
    \label{fig:taxonomy}
\end{figure}

To help show the relation between frequency and generality and characterize the source of vertex sparsity, we illustrate a hierarchical taxonomy of predicates ordered by specificity, following from the theories of natural categories and prototype instances \cite{rosch1975family, rosch1976basic}. We place very general predicate categories at the top of this taxonomy such as ``act'' and ``move,'' with concrete subcategories beneath, and highly specific ones at the bottom, like ``innoculate'' and ``perambulate.'' Rosch et al define their middle ``basic level categories'' for nouns, containing everyday concepts like ``dog'' and ``table,'' which are learned early by humans and are used most commonly among all categories, even by adults \cite{mervis1976relationships}. We assume an analogous basic level in a predicate taxonomy, too, in Figure~\ref{fig:taxonomy}.

There are few general categories at the top and many specific ones at the bottom (e.g., consider the many ways to ``move,'' e.g. ``walk,'' ``sprint,'' ``lunge''). However, since basic level categories are the most frequently used, moving either up or down in the taxonomy accompanies a decrease in usage frequency. Above the basic level, predicates are fewer and more abstract, and can be infelicitous in daily use (e.g. calling a cat a ``mammal'' in Rosch's case or predicates like ``actuate'' in ours). Below, predicates are highly specialized for specific contexts, so there are many more of them, and they are lower-frequency (e.g. ``elongate,'' ``defenestrate''). This is a major source of vertex sparsity.

This asymmetry encourages P-smoothing using an LM (and foreshadows its failure at H-smoothing). A predicate $z$ is likely to be missing from an EG if it is corpus-infrequent, thus likely specific. Randomly sampling another EG predicate $z'$ neighboring $z$ in embedding space, but sampled \textit{proportional} to observed frequencies, is likely to return a predicate of higher frequency, toward the basic level, which is usually higher in the specificity taxonomy. Thus given $z$, a frequency-proportional sample $z'$ is likely to be more general than $z$, usable for P-smoothing to construct a transitive chain.




\section{Experimental Methods}
\label{sec:smoothLM}

In this work we consider Entailment Graphs of typed binary predicates. An EG is defined as $G = (V, E)$, consisting of a set of vertices $V$ of natural language predicates (with argument types in the set $\mathcal{T}$), and directed edges $E$ indicating entailments.

Binary predicates in $V$ have two argument slots labeled with their types. For example, the predicate  $\textsc{travel.to}$(:person, :location) $\in V$, and the types :person, :location $\in \mathcal{T}$. An example entailment is $\textsc{travel.to}$(:person, :location) $\vDash$ $\textsc{arrive.at}$(:person, :location) $\in E$.

Our smoothing method may be applied to any existing EG. In this work we show the complementary benefits of vertex-smoothing with existing methods in improving edge sparsity by comparing two related baseline models, described in \S\ref{ssec:smoothing_exps}. These EGs are learned from the same set of vertices, but are constructed differently so have different edges. The FIGER type system is used for these experiments \cite{ling-figer}, where $|\mathcal{T}| = 49$, and these models typically have up to $|\mathcal{T}|^2 = 49^2$ typed subgraphs $g \in \mathcal{G}$. Typing disambiguates senses of the same predicate, which improves precision of inferences, an observation in NLP tracing back to \citet{yarowsky-1993-one}. For example, $\textsc{run}$(:person, :organization) which is learned in the typed subgraph $g^{(\textit{person-organization})}$ has a different meaning and entailments than $\textsc{run}$(:person, :software). 


\subsection{Nearest Neighbors Search}
\label{ssec:smoothLM:method}

Our method assumes that existing EGs contain enough predicates already present in the graph to enable discovery of suitable replacements for an unseen target predicate, using an LM. For example, in the sports domain, the EG may be missing a rare predicate \textsc{obliterate} but contain similar predicates \textsc{beat} and \textsc{defeat} which can be found as close neighbors in Language Model embedding space. These nearby predicates are expected to have similar semantics (and entailments) to the unseen target predicate, and will thus be suitable replacements. See Figure~\ref{fig:method-illustration} for an illustration. 

We define the smoothed retrieval function $S$, which replaces the typical method for retrieving a target predicate vertex $x$ from a typed subgraph $g^{(t)} = (V^{(t)}, E^{(t)})$, with typing $t \in \{\mathcal{T} \times \mathcal{T}\}$.

Ahead of test-time, for each typed subgraph $g^{(t)}$ we encode the EG predicate vertices $V^{(t)}$ as a matrix $\textbf{V}^{(t)}$. For each predicate $v^{(t)}_i \in V^{(t)}$, we encode $\textbf{v}_i^{(t)}=L(v_i^{(t)})$, a row vector $\textbf{v}^{(t)}_{i} \in \textbf{V}^{(t)}$.

At test-time we encode a corresponding vector for the target predicate $x$, $\textbf{x}=L(x)$. Then $S$ retrieves the K-nearest neighbors of $x$ in $g^{(t)}$:
\begin{align*}
    & S(x, g^{(t)}, K) = \\
    & \{v^{(t)}_{i} \mid v^{(t)}_{i} \in V^{(t)}, \text{ if } \textbf{v}^{(t)}_{i} \in \textit{KNN}(\textbf{x}, \textbf{V}^{(t)}, K)\}
\end{align*}


$L(\cdot)$ is a function which encodes a typed natural language predicate using a pretrained LM. First, a short sentence is constructed from the predicate using the types as generic arguments, and then the sentence is encoded by the LM (see Table~\ref{tab:sents} for examples). We extract the representations of WordPieces corresponding to the predicate, and average them into the resulting predicate vector. In our experiments we use \textbf{RoBERTa} \cite{liu2019roberta} for encoding, a well-tested, off-the-shelf LM of tractable size for running on a single GPU, which has pretrained on 160GB of unlabeled text.


\begin{table}[t]
\centering
\small
\begin{tabular}{@{}ll@{}}
\toprule
$x:$ \relation{(join.1,join.2)\#person\#organization} \\ \quad $\Rightarrow$ ``person \predicatetext{join} organization'' \\[0.2cm]
$x:$ \relation{(give.2,give.to.2)\#medicine\#person} \\ \quad $\Rightarrow$ ``\predicatetext{give} medicine \predicatetext{to} person'' \\[0.2cm]
$x:$ \relation{(export.1,export.to.2)\#location\_1\#location\_2} \\ \quad $\Rightarrow$ ``location\_1 \predicatetext{export to} location\_2'' \\
\bottomrule
\end{tabular}
\caption{A typed predicate $x$ is converted to a sentence (shown) and encoded with an LM by $L(x)$. The final output is the average over \predicatetext{predicate} WordPiece vectors.}
\label{tab:sents}
\end{table}

For the KNN search metric we use Euclidean Distance (L$^2$ norm) from the target vector $\textbf{x}$ to vectors in $\textbf{V}^{(t)}$. We precompute a BallTree using scikit-learn \cite{scikit-learn} which spatially organizes the EG vectors to speed up search from linear in the number of vertices $|V^{(t)}|$ to log $|V^{(t)}|$. 

\subsection{Datasets}
\label{ssec:smoothLM:datasets}

We demonstrate our smoothing method on two explicitly directional datasets, which test both directions of predicate inference, creating a 50\% positive/50\% negative class balance.

\textbf{Levy/Holt}. This dataset \cite{holt_2018, levy-dagan-2016-annotating} has been explored thoroughly in previous work \cite{hosseini2021unsupervised, guillou-etal-2021-blindness, li-etal-2022-cross-lingual, chen-etal-2022-transitivity}. Importantly, it includes inverses for all queries, allowing systematic investigation of directionality, although it contains a high proportion of paraphrases and selection bias artifacts that can be picked up by finetuning in supervised models \cite{li-lm-poor-learner}. We test on the 1,784 questions forming the purely directional subset, which is more challenging.

\textbf{ANT}. This is a new, high-quality dataset improving on Levy/Holt, which tests predicate entailment in the general domain \cite{Ant_2023}. It was created by expert annotation of entailment relations between clusters of predicate paraphrases, expanded automatically using WordNet and other dictionary resources into thousands of test questions of the format ``given [premise], is [hypothesis] true?'' We test on the directional subset of 2,930 questions.

See Table~\ref{tab:datasets} for dataset examples. Each comes preprocessed with argument types from CoreNLP \cite{CoreNLP, finkel_NER}, roughly aligning with EG FIGER types. We use the MoNTEE system \cite{devroe2021modality} to extract CCG-parsed and typed predicate relations ($x$) shown in Table~\ref{tab:sents}, which are used as queries to Entailment Graphs.

\begin{table}[t]
\centering
\small
\begin{tabular}{@{}p{0.9\linewidth}@{}}
\toprule
``The audience applauded the comedian'' $\vDash$ ``The audience observed the comedian'' \\[0.4cm]
``The audience observed the comedian'' $\nvDash$ ``The audience applauded the comedian'' \\
\midrule
``The laptop satisfied the criteria'' $\vDash$	``The laptop was assessed against the criteria'' \\[0.4cm]
``The laptop was assessed against the criteria'' $\nvDash$ ``The laptop satisfied the criteria'' \\
\bottomrule
\end{tabular}
\caption{Example queries, ANT (dev) directional subset.}
\label{tab:datasets}
\end{table}

\subsection{Models}

We smooth two recent Entailment Graphs which previously scored highly amongst unsupervised models on the full Levy/Holt dataset. Importantly, they are constructed from the same set of predicate vertices but have different edges, so we can observe how vertex- and edge-improvements combine.

\textbf{GBL}. The EG of \citet{hosseini2018}, which introduces a ``globalizing'' graph-based method to improve the edges after ``local'' EG learning.

\textbf{CTX}. The state-of-the-art contextualized EG of \citet{hosseini-etal-2021-open-domain}, which improves over GBL edges by augmenting local learning with a contextual link-prediction objective, before globalizing.


\textbf{GBL-P} / \textbf{GBL-H} and \textbf{CTX-P} / \textbf{CTX-H}. We apply an LM separately for both P- and H-smoothing on GBL and CTX. As described earlier, we use the RoBERTa LM \cite{liu2019roberta} to produce embeddings for smoothing the EG.


\textbf{S\&S}. The finetuned RoBERTa model of \citet{schmitt-schutze-2021-language} (discussed in \S\ref{sec:background}). We insert each premise/hypothesis pair into their 5 prompt templates, and take the maximum entailment score as the model prediction for the pair. \citet{li-lm-poor-learner} find that this model has overfit to artifacts present in Levy/Holt, so we compare with it on a different question answering task in \S\ref{sec:qaeval}.

\section{Experiment 1: Entailment Detection}
\label{ssec:smoothing_exps}

We run two experiments on both Levy/Holt and ANT. (1) We apply our unsupervised smoothing to augment the \textit{\textbf{P}remise} of each test entailment, generating $K$ new target premise predicates. Separately, (2) we smooth the \textit{\textbf{H}ypothesis} of each test entailment the same way. For both we try different values of the hyperparameter $K \in$ \{2, 3, 4\}.

\begin{figure}[t]
\centering
\vspace{-0.5cm}
\includegraphics[width=\linewidth]{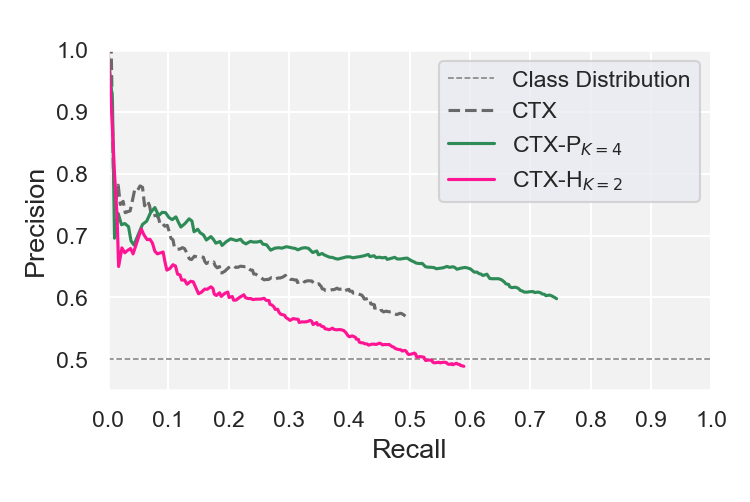}
\vspace{-1cm}
\caption{Experiment 1: Smoothing the CTX EG with an LM on the ANT dataset. P-smoothing improves recall and precision, whereas H-smoothing is detrimental. We try different $K \in \{2,3,4\}$ and show the best $K_{prem}=4$ and $K_{hyp}=2$.}
\label{fig:p_h_lm_ant_ctx}
\end{figure}

\label{ssec:smoothLM:results}

Plots for model performances are shown in Figure~\ref{fig:p_h_lm_ant_ctx}, in which we compare P-smoothing vs. H-smoothing of the CTX graph using $K_{prem}=4$ and $K_{hyp}=2$, chosen for producing the best AUC$_n$ (see Appendix~\ref{apx:hyperparameters} for all results). In Appendix~\ref{apx:gbl} we also show P-smoothing in particular of the CTX graph vs. the GBL graph. For all models (best K selected) on both datasets we show summary statistics in Table~\ref{tab:results}, including \textit{normalized} area under the precision-recall curve (\textbf{AUC$_n$}) and average precision (\textbf{AP}) across the recall range. A sample of model outputs is shown in Table~\ref{tab:examples}.

\citet{li-lm-poor-learner} introduce AUC$_n$, a fair way to compare models which may achieve different maximum recalls. It computes only the area under the precision-recall curve \textit{above} the random-guess baseline for the dataset, so it is highly discerning compared to AUC, which can inflate performance when there is a high random baseline. In our case, the high 50\% random baseline means that AUC$_n$ scores are systematically much smaller than AUC.


\begin{table}[tb]
\footnotesize
\centering
\begin{tabular}{lrrrr}
\toprule
& \multicolumn{2}{c}{\bf ANT} & \multicolumn{2}{c}{\bf Levy/Holt} \\
\cmidrule(lr){2-3} \cmidrule(lr){4-5}
\textbf{Model} & \multicolumn{1}{c}{AUC$_{n}$} & \multicolumn{1}{c}{AP} & \multicolumn{1}{c}{AUC$_{n}$} & \multicolumn{1}{c}{AP} \\
\midrule
GBL & 3.79 & 58.36 & 3.01 & 55.82 \\
GBL-P$_{K=4}$ & \textbf{13.91} & \textbf{64.71} & \textbf{9.95} & \textbf{60.70} \\
GBL-H$_{K=2}$ & 1.41 & 52.57 & 1.09 & 52.05 \\
\midrule
CTX & 15.44 & 65.66 & 9.40 & 60.19 \\
CTX-P$_{K=4}$ & \textbf{25.86} & \textbf{67.47} & \textbf{13.45} & \textbf{60.80} \\
CTX-H$_{K=2}$ & 9.94 & 58.52 & 8.33 & 57.97 \\
\bottomrule
\end{tabular}
\caption{Experiment 1: P- and H-smoothing, compared to unsmoothed models. P-Smoothing with an LM improves AUC$_{norm}$ (AUC$_n$) and Average Precision (AP) in both CTX and GBL models.}
\label{tab:results}
\end{table}

As predicted, our method of selecting nearest-neighbors of a target predicate in an EG using their LM embedding distance has different behavior for P-smoothing than H-smoothing. We observe that P-smoothing with an LM is very beneficial to both the recall and precision of both Entailment Graphs it is applied to, with a slight advantage in AUC$_n$ to higher values of K. When applied to the SOTA model CTX on the ANT dataset, our smoothing method increases maximum recall by 25.1 absolute percentage points (pp) to 74.3\% while increasing average precision from 65.66\% to 67.47\%. On Levy/Holt we increase maximum recall by 16.3 absolute pp to 62.7\% while slightly raising average precision. However, H-smoothing with the LM is highly detrimental: despite improving recall, average precision on ANT is cut to 58.52\%, and the lowest confidence predictions are at random chance (50\% precision). 

We also note that P-smoothing greatly improves recall and precision when applied to \textit{both} GBL and CTX graphs. This shows the complementary nature of improving vertex sparsity with improving edge sparsity in EGs: these techniques improve different aspects, which can be applied together. Since effects are similar for both EGs, from now on we show results only for CTX, and report additional results for the weaker GBL in Appendix~\ref{apx:gbl}.

\begin{table}[tb]
\centering
\small
\begin{tabular}{@{}>{\raggedright\arraybackslash}p{0.53\linewidth}>{\raggedright\arraybackslash}p{0.33\linewidth}@{}}
\toprule
\multicolumn{1}{>{\centering\arraybackslash}p{0.38\linewidth}}{\textbf{Predicate Missing\newline from EG}} & \multicolumn{1}{>{\centering\arraybackslash}p{0.35\linewidth}}{\textbf{Nearest Neighbors \newline by Embedding Dist.}} \\
\midrule
\textsc{discredit}(:person, :thing) & \textsc{probe}, \textsc{accuse} \\[0.15cm] 

\textsc{crack.up.at}(:person, :written\_work) & \textsc{make.joke.at}, \textsc{yell.at} \\[0.6cm]

\textsc{minimize}(:organization, :thing) & \textsc{soften},  \textsc{evade} \\[0.2cm] 

\textsc{rebuke}(:person, :person) & \textsc{oppose}, \textsc{remind} \\ 
\bottomrule
\end{tabular}
\caption{Experiment 1: Sample of CTX outputs on ANT. A target \textsc{predicate}(type1, type2) that is missing from CTX is closest in LM embedding space to K=2 CTX predicates, which are more semantically general.}
\label{tab:examples}
\end{table}

\section{Experiment 2: Question Answering}
\label{sec:qaeval}

We now experiment with LM smoothing in application on an applied task. We test on the Boolean Open QA task, BoOQA \cite{li-lm-poor-learner}, in which models answer true/false questions about entities mentioned in news articles from multiple sources. BoOQA questions are chosen to be adversarial to simple similarity baselines, and EGs have proven useful by using directional reasoning. 

\subsection{Boolean Open-Domain QA}
BoOQA is a task over open domain news articles, with questions formed by extracting triples of (entity, relation, entity), in the format ``is it true that \texttt{<triple>}?'' \textit{Context statements} are other triples sourced from the articles concerning the same question entities, and the task is to compare each context statement with the question itself. If any context statement entails the question by means of its relation, the question can be labeled ``true,'' otherwise ``false.'' BoOQA also contains false questions derived from true ones, so models must decide carefully what is supported by evidence and what isn't.

We address vertex sparsity in a natural setting, so we relax the original entity restriction of \citet{li-lm-poor-learner}: instead of sampling questions about frequently-mentioned entities (which always have many context statements to decide from), we increase the challenge by sampling from the natural distribution of entities, regardless of popularity.





\subsection{Results Across Context Sizes}

\begin{table}[]
    \small
    \centering
    \begin{tabular}{l|c|cc|c}
    \toprule
        Context Size & CTX & CTX-P & CTX-H & S\&S \\
        \midrule
        $[2, 5)$ & 20.05 & \textbf{20.66} & 19.07 & 17.00 \\
        $[5, 10)$ & 29.13 & \textbf{29.17} & 29.01 & 23.05 \\
        $[10, 15)$ & \textbf{32.32} & 32.31 & 32.25 & 24.98 \\
        15+ & \textbf{36.58} & 36.57 & 36.51 & 26.13 \\
        \midrule
        All Questions & 21.26 & \textbf{21.74} & 20.64 & 16.99 \\
        \bottomrule
    \end{tabular}
    \caption{Experiment 2: Effect of P- and H-smoothing vs. baseline CTX and S\&S across context sizes (AUC$_n$ is reported). P-smoothing is useful on CTX when fewer context statements are available.}
    \label{tab:qa_freqbands}
    \vspace{-0.15in}
\end{table}

Results corroborate the earlier tests: P-smoothing improves AUC$_n$ from 21.26\% to 21.74\% over all questions, while H-smoothing worsens to 20.64\% (as in \S\ref{sec:smoothLM}, AUC$_n$ is systematically lower than AUC). We also outperform \citet{schmitt-schutze-2021-language}, our most direct competition which uses a tractable-size LM. Despite facility to encode any predicate, it lacks directional precision useful for this task.

To demonstrate when smoothing an EG is helpful, we further analyze the effect on different \emph{context size bands}. For each question, we count the number of context sentences available to answer it; questions are bucketed into bands of $[2, 5)$, $[5, 10)$, $[10, 15)$, 15+. From the overall dataset we sample approximately 55,000 questions per context size band (see Appendix~\ref{apx:booqa_dataset_sizes} for exact counts). On each band we compare an unsmoothed model with P-smoothing and H-smoothing, and we report results in Table~\ref{tab:qa_freqbands}.

The benefit of P-smoothing is greatest in the lowest band $f<5$, and diminishes in higher bands. This is because in the lower bands there are fewer context statements which may be used to answer the question, increasing difficulty. Here the EGs are more prone to sparsity, because missing even a few context predicates devastates its chance to answer the question. In fact, the proportion of questions for which all context relations are missing from the EG is 1.5\% for $f>15$, but 32.7\% for $f<5$.

\section{Experiment 3: P and H with~WordNet}
\label{ssec:wordnet}

LM P-smoothing works well, but not H-smoothing. We now show controlled experiments using WordNet relations \cite{wordnet} to confirm this is due to semantic generalization (in line with our theory in \S\ref{sec:transitive_chaining}). We show by constructing a transitive chain using WordNet hyponyms that Hypothesis smoothing is possible in principle, without claiming that it provides a practical alternative to an LM.


\begin{figure*}[h]
  \includegraphics[width=\linewidth]{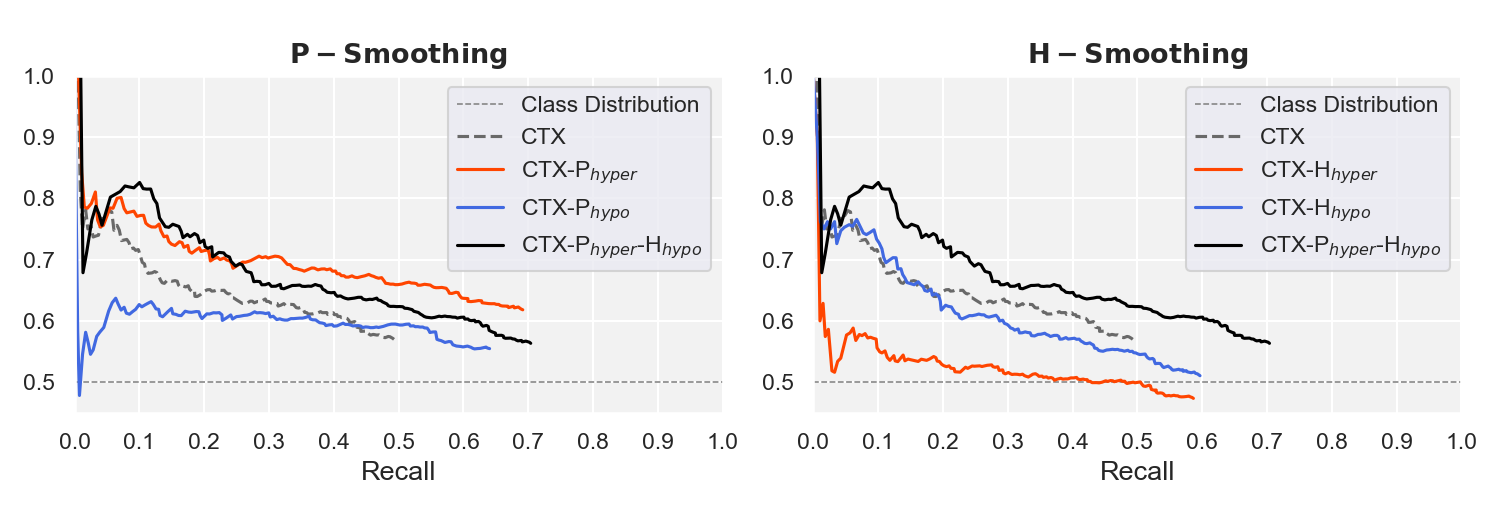}
  \vspace{-1cm}
  \caption{Experiment 3: Comparing WordNet relations used in smoothing P(remise), H(ypothesis), and P+H, with CTX graphs on the ANT dataset. Hypernyms are useful for P-smoothing, and hyponyms less so for H-smoothing.}
  \label{fig:ctx_wn_comparison}
  \vspace{-0.3cm}
\end{figure*}


\subsection{Controlled Search with WordNet}

We re-run the \S\ref{sec:smoothLM} experiment by smoothing the CTX model on the ANT dataset (GBL in Appendix~\ref{apx:gbl}). However, the target premise or hypothesis is now approximated without the LM. Instead, we generate replacements using two WordNet relations.\footnote{WN was partly used in ANT's construction, so this result explains the LM effect, rather than offering a practical model.}

In this test, we choose specific WordNet lexical relations as instances of entailment, then generate smoothing predictions from the WN database. The \textbf{hyponymy} relation is used for specialization and \textbf{hypernymy} for generalization, and these are compared for both P- and H-smoothing. 

To illustrate, if smoothing by specializing, given a predicate ``receive from,'' we retrieve WN hypernyms like ``inherit from.'' We do this by querying WN for relations of the predicate head word. We use results from the first word sense to replace the query word. E.g., from \texttt{(receive.2,receive.from.2)} the WN query \textit{hyponym}(``receive'') $\Rightarrow$ ``inherit'' generates \texttt{(inherit.2,inherit.from.2)}.

\subsection{Results}

Results are shown in Figure~\ref{fig:ctx_wn_comparison}. 
Importantly, from these plots a switch in performance is observed between the application of hypernyms and hyponyms when used for P- and H-smoothing on CTX (similar results for GBL, see Appendix~\ref{apx:gbl}).
It is clear that generalizing the premise using hypernyms is highly effective in terms of recall and precision, but specializing with hyponyms is extremely damaging to precision. For the hypothesis, the reverse is true: generalizing with hypernyms worsens performance, but specializing with hyponyms can lead to some performance gains (when used with P-smoothing, see discussion below). We also tested Levy/Holt and see a similar trend.

These results nearly replicate the behavior of the LM-smoother in \S\ref{sec:smoothLM}, verifying that nearest neighbor search in LM embedding space has a semantically generalizing effect suitable for P-smoothing. Table~\ref{tab:examples} shows examples of generalized predictions.

Finally, we note P-smoothing with WordNet performs similarly to the LM in this ``laboratory'' setting (see Appendix~\ref{apx:p_lm_wn}), but an LM smoother is still preferable due to being fully automatic and open-domain, handling new words, misspellings, etc.

\subsection{Discussion}

We note two phenomena of interest. (1) For both CTX and GBL, performance is boosted in the low-recall/high-precision range when using both optimal smoothers ($P_{hyper} + H_{hypo}$), higher than using either smoother individually. (2) Additionally, $H_{hypo}$ is the better $H$ smoother tested, though it appears unreliable on its own without $P$ smoothing: $H_{hypo}$ is not useful for smoothing CTX, but it does improve the weaker GBL, see Appendix~\ref{apx:gbl}.

Both of these phenomena are likely related to data frequency. Generalized hypernyms such as \textsc{beat} and \textsc{use} are quite common in training data, and therefore have more learned edges in the EG with high quality edge weights. However, specialized hyponyms like \textsc{elongate} can be extremely sparse in training data, leading to poorer learned representations and fewer edges. Phenomenon (1) shows that using a frequently-occurring smoothed premise of high quality yields better odds of finding an edge to a smoothed hypothesis, leading to some performance gains over either smoother individually. Phenomenon (2) suggests that H-smoothing may be naturally more difficult than P-smoothing, and less stable due to sparsity of hyponyms (specializations) in corpora. If a hypothesis $h$ is missing from the EG (meaning it wasn't seen in training) then deriving a candidate for replacement $h'$ specialized from $h$ will also be unlikely to occur in training, thus even if found in the EG it may have few or poorly learned edges. Though it can be beneficial to precision, natural data sparsity makes H-smoothing fundamentally harder.

\section{Conclusion}

We introduce a theory for optimal smoothing of a symbolic model of language inference like an Entailment Graph, which solves the problem of vertex sparsity in EGs by constructing transitive inference chains. Further, we show an unsupervised, open-domain method of P-smoothing by approximating premises missing from an EG using Language Model embeddings, which improves both recall and precision on two difficult directional entailment datasets. We also test the method on a QA task, where we show the most benefit in difficult scenarios where limited context information is available. Our method is low-compute, combining an existing EG with a pretrained LM of tractable size for a single GPU, and it improves over two low-compute baselines: a SOTA EG and a finetuned RoBERTa-based prompting model.

We also demonstrate our theory of optimal smoothing by directing the search for predictions using WordNet relations, without an LM. Our experiments replicate the behavior of the LM-based smoother, offering an explanation for why LM embeddings are useful for P-smoothing, but not H-smoothing, in terms of the semantic generalizing effect when searching a neighborhood in embedding space.

\section*{Limitations}

In this work we present a simple ``graph smoothing'' method which leverages the natural structure in LM embedding space to find approximations of predicates missing from the EG, a major source of error. Nearest neighbors search within LM embedding space is biased toward returning predicates that are more semantically general, which is helpful for P-smoothing.

However, generalizing is detrimental to H-smoothing, which requires specialization. While we show a proof of specialization and empirical evidence using WordNet, solving H-smoothing in an open domain using an unsupervised model such as a Language Model is left open in this work. It is likely that H-smoothing is a more difficult task than P-smoothing due to natural data sparsity as discussed in the paper. If a hypothesis is missing from the EG, it is likely to be a corpus-infrequent predicate, and specializing it will yield other predicates of low frequency, yielding poor odds of recovery.

Further, using a sub-symbolic LM encoder theoretically enables inference using any premise predicate, but we are still restricted to choosing approximations from the predicate vocabulary of the EG. If the vocabulary is not suitable e.g. for a new target genre/domain, \citet{hosseini-etal-2021-open-domain} show that EG learning may be scaled up easily, which may provide a sufficiently scoped vocabulary for any application, but exploration is left to future work.

Finally, our work is demonstrated only on the English language. However, we expect this method should succeed with arbitrary natural languages. \citet{li-etal-2022-cross-lingual} demonstrate that learning Entailment Graphs in Chinese can be done using the same process as English, and our technique leverages a simple fundamental property of Language Models stemming from the natural Zipfian distribution of predicates in corpora, across languages.

\section*{Ethical Considerations}

This work is designed to extend the capabilities of Entailment Graphs, which are general-purpose structures of meaning postulates. These can be applied most readily to question answering applications, but they can also be used for other NLU or NLI tasks. As an unsupervised, corpus-based learning algorithm, we believe that EGs could be susceptible to learning biases in human beliefs present in corpora, but this algorithm is most sensitive to widely repeated statements, which may be easier to detect in data cleaning than uncommon statements. We believe there is no immediate risk in basic question answering when using EGs that are trained on published news articles, as shown in this work, because the training data is professionally edited to a standard. However, models for general language understanding like an EG may be used for many purposes beyond this.

\section*{Acknowledgements}

This research was supported by ERC Advanced Fellowship GA 742137 SEMANTAX, the University of Edinburgh Huawei Laboratory, and a Google Faculty Award.

\bibliography{custom}
\bibliographystyle{acl_natbib}

\appendix

\section{Hyperparameter Search}
\label{apx:hyperparameters}

\begin{figure}[h!]
\begin{center}
\includegraphics[width=\linewidth]{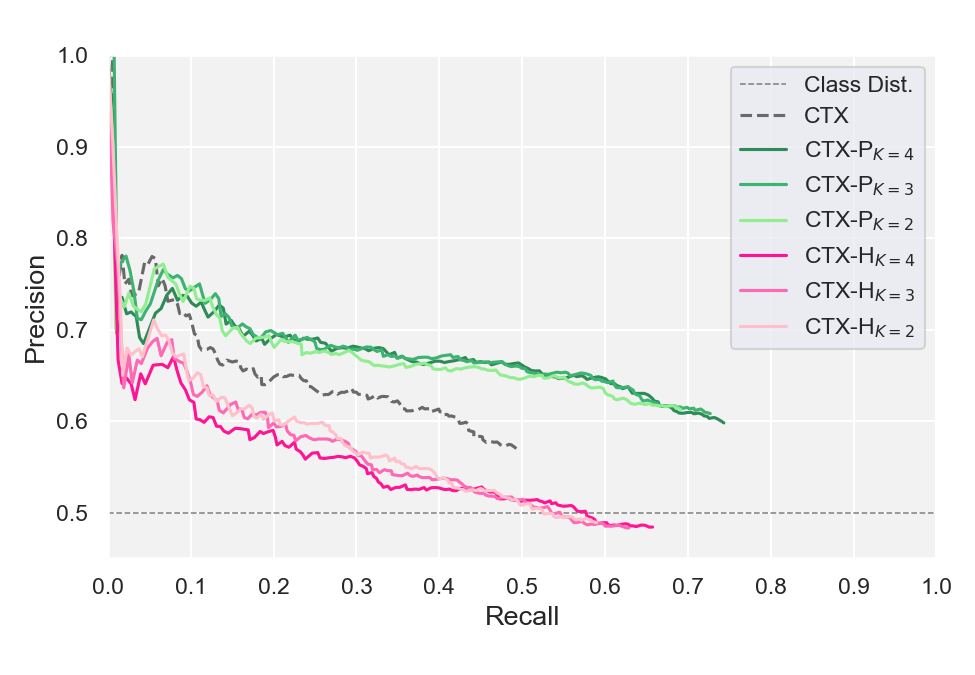}
\end{center}
\caption{Experiment 1: LM smoothing on the ANT dataset. Comparison of P- and H-smoothing CTX with different $K_{prem}$ and $K_{hyp}$ from choices \{2, 3, 4\}. \textit{Higher values of K are shown more darkly.}}
\label{fig:hyperparameters}
\end{figure}

In \S\ref{ssec:smoothing_exps} we test three values for hyperparameters $K_{prem}$ and $K_{hyp}$, each from choices \{2, 3, 4\}. Figure~\ref{fig:hyperparameters} shows all smoothing combinations. We select $K_{prem}=4$ and $K_{hyp}=2$ in the main experiments due to having the highest AUC$_n$ values for P- and H-smoothing, respectively. We highlight a few trends. (1) higher $K_{prem}$ appears better (most notably, $K_{prem}=4$ yields slightly better recall than $K_{prem}=2$), though it has diminishing returns. (2) lower $K_{hyp}$ is better, because H-smoothing using an LM is actively harmful ($K_{hyp}=0$, an unsmoothed EG, would ``perform'' better in practice!).

\section{The GBL Entailment Graph}
\label{apx:gbl}
\begin{figure}[h!]
\begin{center}
\includegraphics[width=\linewidth]{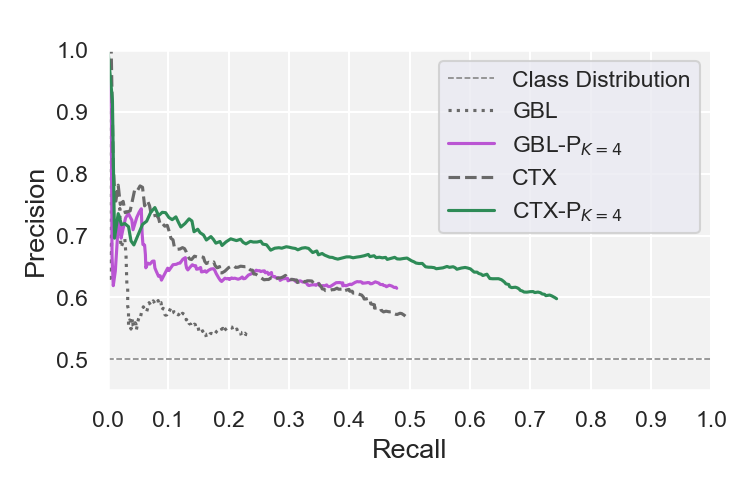}
\end{center}
\vspace{-0.75cm}
\caption{Experiment 1: LM smoothing on the ANT dataset. Comparison of P-smoothing GBL and CTX with optimal K=4.}
\label{fig:p_h_lm_ant_gbl}
\end{figure}

We test the older GBL graph \cite{hosseini2018} on the ANT dataset. Results confirm findings on the newer CTX \cite{hosseini-etal-2021-open-domain}. Figure~\ref{fig:p_h_lm_ant_gbl} shows results for the experiment in \S\ref{sec:smoothLM} but comparing P-smoothing with LM predictions for the CTX and GBL graphs. We note that base CTX performs much better than GBL, and that P-smoothing with an LM improves both GBL and CTX.

\begin{figure*}[t]
  \includegraphics[width=\linewidth]{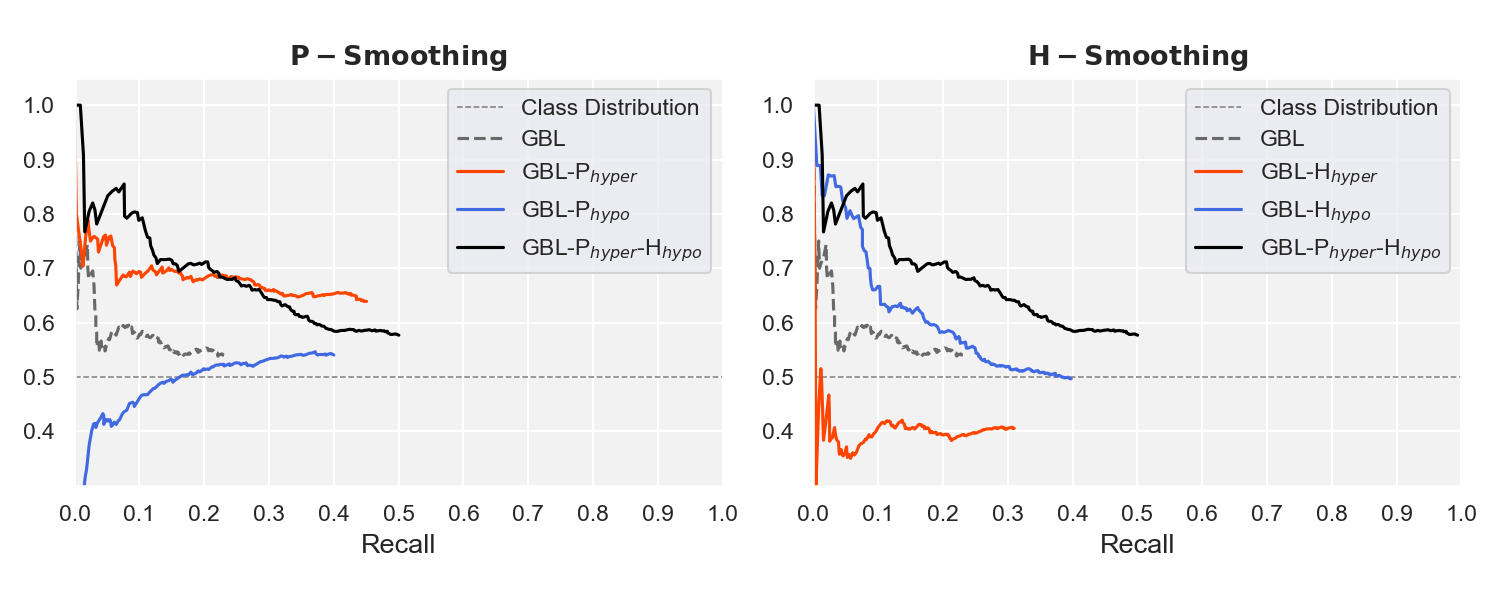}
  \vspace{-1cm}
  \caption{Experiment 3: WordNet relations used to smooth P(remise), H(ypothesis), and P+H, with the Entailment Graph GBL on the ANT dataset. Hypernyms are useful for P-smoothing, and hyponyms for H-smoothing.}
  \label{fig:gbl_wn_comparison}
\end{figure*}

Figure~\ref{fig:gbl_wn_comparison} shows results for the experiment in \S\ref{ssec:wordnet} of smoothing an EG using WordNet relations, but we now show smoothing the older GBL graph. We observe similar results as with CTX: there is noticeable improvement over the base EG when smoothing either premises with hypernyms, hypotheses with hyponyms (stronger than when applied to CTX), or both combined.

\section{BoOQA Context Size Bands}
\label{apx:booqa_dataset_sizes}

In the QA task a model must try to draw an inference from any context statement (premises) to infer the validity of the question (hypothesis). Any model is less likely to find an entailment when there are few premises, but symbolic EGs are especially prone because missing premises means even fewer chances to find an entailment. From the original dataset, we sample approximately 55,000 questions for each context size band, including 55,000 questions from the natural distribution, with no context limitation (``All Questions''). Sample sizes are shown in Table~\ref{tab:booqa_sample_sizes}.

\begin{table}[h]
    \centering
    \begin{tabular}{ll}
        \toprule
        $[2, 5)$ & 56,390 \\
        $[5, 10)$ & 56,425 \\
        $[10, 15)$ & 54,778 \\
        15+ & 54,926 \\
        \midrule
        All Questions & 56,494 \\
        \bottomrule
    \end{tabular}
    \caption{Experiment 2: Sample sizes for context bands on the QA task.}
    \label{tab:booqa_sample_sizes}
    \vspace{-0.2in}
\end{table}

\section{P-Smoothing: LM vs. WordNet}
\label{apx:p_lm_wn}
\begin{figure}[h]
  \includegraphics[width=\linewidth]{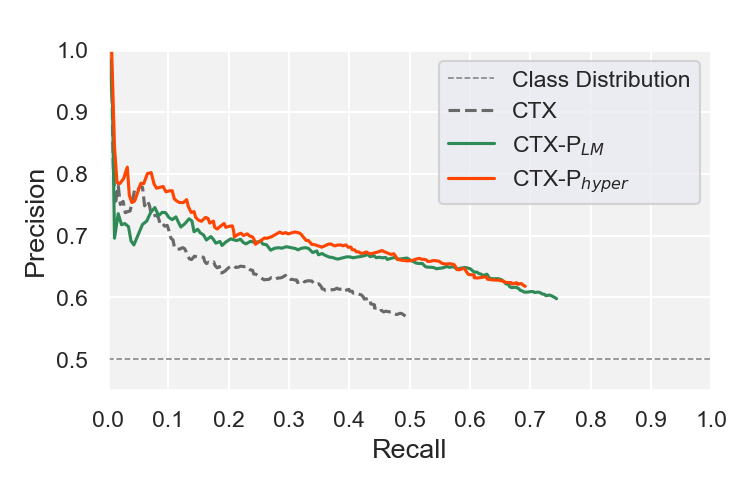}
  \vspace{-1cm}
  \caption{Comparison of P-smoothing methods on ANT: LM-based smoother performs similarly to WordNet hypernym relations on the Entailment Graph CTX. }
  \label{fig:p_lm_wn_comparison}
  \vspace{-0.2in}
\end{figure}

In Figure~\ref{fig:p_lm_wn_comparison} we show a comparison of P-smoothing between the LM (CTX-P$_{LM}$ AUC$_n=25.86$) and WordNet (CTX-P$_{hyper}$ AUC$_n=27.39$) on the ANT dataset. We note that although WordNet performs within about 1.5\% of the LM smoother in this ``laboratory'' experiment, we believe the LM-smoother is preferable in use, because it is fully automatic to learn and apply, and because it encodes an open domain of predicates, which may include new words, misspellings, etc. that WordNet cannot handle.

\end{document}